\documentclass[11pt,a4paper]{article}
\usepackage[hyperref]{naaclhlt2018}
\usepackage{times}
\usepackage{latexsym}
\usepackage{todonotes}
\usepackage{amssymb}
\usepackage{amsfonts}
\usepackage{url}
\usepackage{multirow} 
\usepackage{graphicx}

\usepackage{amsmath}
\usepackage[utf8]{inputenc}
\usepackage{tipa}
\usepackage{newunicodechar}
\newunicodechar{ə}{@}
\newunicodechar{ɪ}{I}
\newunicodechar{ɛ}{E}

\aclfinalcopy 
\def\aclpaperid{551} 

\title{Modeling Noisiness to Recognize Named Entities using Multitask Neural Networks on Social Media}

\author{Gustavo Aguilar$^\star$ A. Pastor L\'{o}pez-Monroy$^\star$  Fabio A. González$^\dagger$ Thamar Solorio$^\star$ \\
$^\star$Department of Computer Science, University of Houston\\
$^\dagger$Systems and Computer Engineering Department, Universidad Nacional de Colombia\\
{\tt \{gaguilaralas, alopezmonroy\}@uh.edu} \\
{\tt fagonzalezo@unal.edu.co, solorio@cs.uh.edu}
}

\date{}

\begin{document}
\maketitle

\begin{abstract}
Recognizing named entities in a document is a key task in many NLP applications. Although current state-of-the-art approaches to this task reach a high performance on clean text (e.g. newswire genres), those algorithms dramatically degrade when they are moved to noisy environments such as social media domains. We present two systems that address the challenges of processing social media data using character-level phonetics and phonology, word embeddings, and Part-of-Speech tags as features. The first model is a multitask end-to-end Bidirectional Long Short-Term Memory (BLSTM)-Conditional Random Field (CRF) network whose output layer contains two CRF classifiers. The second model uses a multitask BLSTM network as feature extractor that transfers the learning to a CRF classifier for the final prediction. Our systems outperform the current F1 scores of the state of the art on the Workshop on Noisy User-generated Text 2017 dataset by 2.45\% and 3.69\%, establishing a more suitable approach for social media environments. 
\end{abstract}

\section{Introduction}

One of the core tasks in Natural Language Processing (NLP) is Named Entity Recognition (NER). NER is a sequence tagging task that consists in selecting the words that describe entities and recognizing their types (e.g., a person, location, company, etc.).
Figure \ref{f:example} shows examples of sentences from different domains that contain named entities. Recognizing entities in running text is typically one of the first tasks in the pipeline of many NLP applications, including machine translation, summarization, sentiment analysis, and question answering. 

Traditional machine learning systems have proven to be effective in formal text, where grammatical errors are minimal and writers stick to the rules of the written language \citep{Florian:2003:NER, Chieu:2003:NER}. However, those traditional systems dramatically fail on informal text, where improper grammatical structures, spelling inconsistencies, and slang vocabulary prevail \citep{Ritter:2011:NER:2145432.2145595}. For instance, Table \ref{t:snapshot} shows a snapshot of NER systems' performance during the last years, where the results drop from 96.49\% to 41.86\% on the F1 metric as we move from formal to informal text. Although the results are not directly comparable because they consider different conditions and challenges, they serve as strong evidence that the NER task in social media is far from being solved.

\begin{figure}
\renewcommand{\arraystretch}{1.1}
\centering
\small
\begin{tabular}{cc}
\begin{tabular}{|l|}
\hline
{\bf CoNLL 2003 } \\\hline
[\textbf{\textit{Spanish}}]\textsubscript{MISC} Farm Minister [\textbf{\textit{Loyola de Palacio}}]\textsubscript{PER} \\
had earlier accused [\textbf{\textit{Fischler}}]\textsubscript{PER} at an [\textbf{\textit{EU}}]\textsubscript{ORG} \\
farm ministers ' meeting of causing unjustified alarm \\ 
through " dangerous generalisation . " \\
\hline
\end{tabular} 
\\
\\
\begin{tabular}{|l|l|}
\hline
{\bf WNUT 2017, Twitter domain } \\\hline
been listenin to [\textbf{\textit{trey}}]\textsubscript{PER} alllll week ... can \\ 
u luv someone u never met ?? bcuz i think \\
im in luv yeeuuuuppp !!! \\
\hline
\end{tabular}
\end{tabular}
\caption{Examples from the CoNLL 2003 and the WNUT 2017 datasets. The noise from the WNUT dataset makes a clear difference from one text to the other, establishing new challenges to the current state-of-the-art systems on formal text. The words in bold are grouped to described the entities.}
\label{f:example}
\end{figure}

\begin{table*}[t!]
\centering
\begin{tabular}{lllll}
  \hline
  Organizer & Competition & Domain & F1 & Classes \\
  \hline
  \citet{GrishmanAndSundheim:96} & MUC-6 & Newswire  & 96.49\% & 2 \\
  \citet{TjongKimSang-DeMeulder:03} & CoNLL & Newswire  & 88.76\% & 4 \\
  \citet{StraussEtAl:16} & WNUT  & Twitter & 52.41\% & 10 \\
  \citet{DerczynskiEtAl:17} & WNUT  & SM domains & 41.86\% & 6 \\
  \hline
\end{tabular}
\caption{Results on different NER shared tasks. The performance degrades as the systems are moved to social media (SM) environments. The last row considers multiple SM domains, such as Twitter, YouTube, Reddit, and StackExchange.}
\label{t:snapshot}
\end{table*}

Recently, researchers have approached NER using different neural network architectures. For instance, \citet{DBLP:journals/corr/ChiuN15} proposed a neural model using Convolutional Neural Networks (CNN) for characters and a bidirectional Long Short Term Memory (LSTM) for words. Their model learned from word embeddings, capitalization, and lexicon features. On a slightly different approach, \citet{DBLP:journals/corr/LampleBSKD16} used a BLSTM with a CRF at the output layer, removing the dependencies on external resources. Moreover, \citet{MaAndHovy:16} proposed an end-to-end BLSTM-CNN-CRF network, whose loss function is based on the maximum log-likelihood estimation of the CRF. These architectures were benchmarked on the standard CoNLL 2003 dataset~\citep{TjongKimSang-DeMeulder:03}. Although most of the work has focused on formal datasets, similar approaches have been evaluated on SM domains~\citep{StraussEtAl:16, DerczynskiEtAl:17}. In the Workshop on Noisy User-generated Text (WNUT) 2016, \citet{LimsopathamAndCollier:16}, the winners of the NER shared task, used a BLSTM-CRF model that induced features from an orthographic representation of the text. Later, in the WNUT 2017 shared task, the best performing system used a multitask network that transferred the learning to a CRF classifier for the final prediction~\citep{AguilarEtAl:17}.

In this work we focus on addressing the challenges of the NER task found in social media environments. We propose that what is traditionally categorized as noise (i.e., misspellings, inconsistent orthography, emerging abbreviations, and slang) should be modeled \textit{as is} since it is an inherent characteristic of SM text. Specifically, the proposed models attempt to address i) misspellings using subword level representations, ii) grammatical mistakes with SM-oriented Part-of-Speech tags \citep{owoputi2013improved}, iii) sound-driven text with phonetic and phonological features \citep{bharadwaj-EtAl:2016:EMNLP2016}, and iv) the intrinsic skewness of NER datasets by applying class weights. It is worth noting that our models do not rely on capitalization or any external resources such as gazetteers. The reasons are that capitalization is arbitrarily used on SM environments, and gazetteers are expensive resources to develop for a scenario where novel entities constantly and rapidly emerge \citep{DerczynskiEtAl:17, AugensteinEtAl:17}.

Based on our experiments, we have seen that a multitask variation of the proposed networks improves the results over a single-task network. Additionally, this multitask version, paired with phonetic and phonological features, outperforms previous state-of-the-art results on the WNUT 2017 dataset, and the same models obtain reasonable results with respect to the state of the art on the CoNLL 2003 dataset \citep{TjongKimSang-DeMeulder:03}.

The rest of the paper is organized as follows: \S \ref{sec:methodology} presents the proposed features, the formal description of the models, and the implementation details. \S \ref{sec:dataset} describes the datasets and their challenges. On \S \ref{sec:experiments}, we show the evaluation process of our models and the results. We explain the performance of the models on \S \ref{sec:analysis}. \S \ref{sec:relatedwork} describes related work and, finally, we draw conclusions on \S \ref{sec:conclusion}.

\section{Methods}
\label{sec:methodology}

Our methods are based on two main strategies: i) a representation of the input text using complementary features that are more suitable to social media environments, and ii) a fusion of these features by using a multitask neural network model whose main goal is to learn how entities are contextualized with and without the entity type information.

\subsection{Feature representation}

\noindent\textbf{Semantic features}. Semantic features play a crucial role in our pipeline as they provide contextual information to the model. This information allows the model to infer the presence of entities as well as the entity types. We use the pretrained word embedding model provided by \citet{godin2015multimedia}. This model has been trained on 1 million tweets (roughly 1\% of the tweets in a year) with the skipgram algorithm. We take advantage of this resource as it easily adapts to other SM environments besides Twitter \cite{AguilarEtAl:17}. 

\noindent\textbf{Syntactic features}. Syntactic features help the models deal with word disambiguation based on the grammatical role that the words play on a sentence. That is, a word that can be a verb or a noun in different scenarios may conflict with the interpretations of the models; however, by providing syntactical information the models can improve their decisions. We capture grammatical patterns using the Part-of-Speech (POS) tagger provided by \citet{owoputi2013improved}. This POS tagger has custom labels that are suitable to SM data (i.e., the tagger considers emojis, hashtags, URLs and others). 

\begin{table}
\renewcommand{\arraystretch}{1.1}
\centering 
\small
\begin{tabular}{cc}
\begin{tabular}{|l|l|} \hline
\bf{Sentence} & \bf{IPA} \\\hline
u hav to b KIDDDDING me & /{\textipa{ju hæv tə bi kɪdɪŋ mi}}/ \\ 
you have to be kidding me & /{\textipa{ju hæv tə bi kɪdɪŋ mi}}/ \\\hline
\end{tabular}
\end{tabular}
\caption{Examples of both noisy and normalized text. In both cases, the mappings to the International Phonetic Alphabet (IPA) are the same.}
\label{t:phoneme_example}
\end{table}

\noindent\textbf{Phonetic and phonological features}. We also consider the phonetic and phonological aspects of the data at the character level. In Table~\ref{t:phoneme_example} we show an example of two phrases: the first sentence is taken from SM, and the second one is its normalized representation. Even though the spellings of both phrases are significantly different, by using the phonological (articulatory) aspects of those phrases it is possible to map them to the same phonetic representation. In other words, our assumption is that social media writers heavily rely on the way that words sound while they write. We use the Epitran\footnote{https://github.com/dmort27/epitran} library \citep{bharadwaj-EtAl:2016:EMNLP2016}, which transliterates graphemes to phonemes with the International Phonetic Alphabet (IPA). In addition to the IPA phonemes, we also use the phonological (articulatory) features generated by the PanPhon\footnote{https://github.com/dmort27/panphon} library \citep{mortensen-EtAl:2016:COLING}. These features provide articulatory information such as the way the mouth and nasal areas are involved in the elaboration of sounds while people speak.

\subsection{Models}

\begin{figure}
\centering
\includegraphics[width=\linewidth,height=6cm]{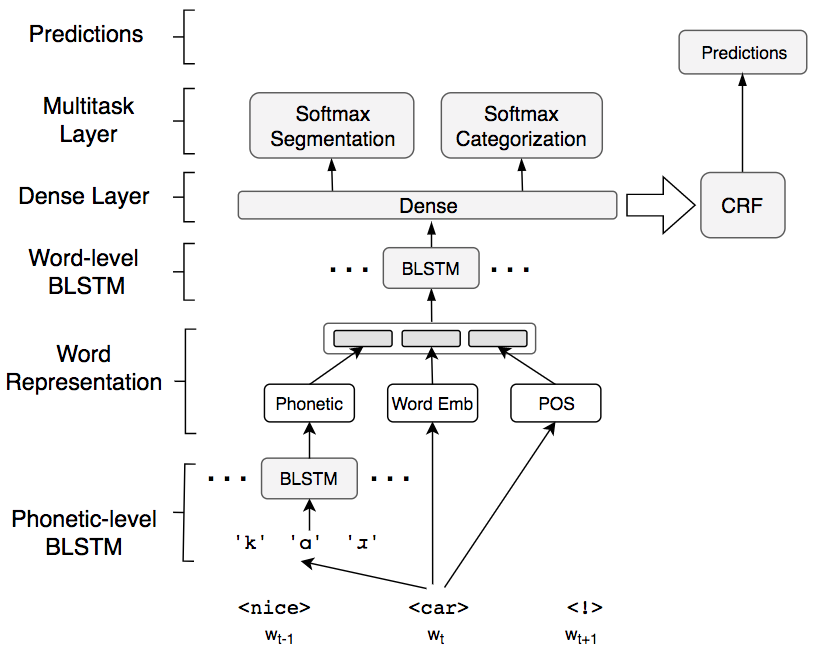}
\caption{ This is a stacked model that uses a network as feature extractor, and then it transfers the learning to a CRF classifier. The network uses multitask learning to capture the features. }
\label{f:system2}
\end{figure}

We have experimented with two models. In the first one, we use an end-to-end BLSTM-CRF network with a multitask output layer comprised of one CRF per task, similar to \citet{DBLP:journals/corr/YangSC16}. In the second one, we define a stacked model that is based on two phases: i) a multitask neural network and ii) a CRF classifier. In the first phase, the network acts as a feature extractor, and then, for the second phase, it transfers the learning to a CRF classifier for the final predictions (see Figure \ref{f:system2}). In both cases, the multitask layer is defined with the following two tasks: 

\begin{itemize}
\item \textbf{Segmentation}. This task focuses on the Begin-Inside-Outside (BIO scheme) level of the tokens. That is, for a given NE, the model has to predict whether a word is B, I, or O regardless of the entity type. The idea is to let the models learn how entities are treated in general, rather than associating the types to certain contexts. This task acts as a regularizer of the primary task to prevent overfitting.
\item \textbf{Categorization}. In this case, the models have to predict the types of the entities along with the BIO scheme (e.g., B-person, I-person, etc.), which represent the final labels.
\end{itemize}

We formalize the definitions of our models as follows: let $X = [x_1, x_2, ..., x_n]$ be a sample sentence where $x_i$ is the $i^{th}$ word in the sequence. Then, let $\alpha:V_x\rightarrow \mathbb{R}^{dim_x}$ be a word embedding, and let $\mathbf{x} = [\alpha(x_1), \ldots, \alpha(x_n)]$ be the word embedding matrix for the sample sentence such that $V_x$ is the vocabulary and $dim_x$ is the dimension of the embedding space. Similarly, let $\beta : V_p \rightarrow  \mathbb{R}^{dim_p}$ be the POS tag embedding, and let $\mathbf{p} =  [\beta(p_1), \ldots, \beta(p_n)]$ be the POS tag embedding matrix for the sample sentence such that $V_p$ is the set of Part-of-Speech tags and  $dim_p$ is the dimension of the embedding space. Notice that the POS tag embedding matrix $\mathbf{p} $ is learned during training. Also, let $Q = [q_1, q_2, ..., q_m]$ be the phonetic letters of a word; let  $\gamma : V_q \rightarrow \mathbb{R}^{|V_q|+dim_{PanPhon}}$ be  an embedding that maps each phonetic character to a one-hot vector of the International Phonetic Alphabet ($V_q$) concatenated with the 21 ($dim_{PanPhon}$) phonological features of the PanPhon library (tongue position, movement of lips, etc.) \citep{bharadwaj-EtAl:2016:EMNLP2016}; and let $\mathbf{q} = [\gamma(q_1), ..., \gamma(q_m)]$ be the matrix representation of the word-level phonetics and phonology.

We first apply an LSTM \citep{Hochreiter:1997:LSM:1246443.1246450} to the $\mathbf{q}$ matrix on forward and backward directions. Then we concatenate the output from both directions:
\begin{equation*} \label{eq:blstm}
\begin{split}
\overrightarrow{\mathbf{h}} = &~\textrm{LSTM}(\{\mathbf{q}_1, \mathbf{q}_2, ..., \mathbf{q}_m\}) \\
\overleftarrow{\mathbf{h}} = &~\textrm{LSTM}(\{\mathbf{q}_m, \mathbf{q}_{m-1}, ..., \mathbf{q}_1\}) \\
\mathbf{h} = &~[\overrightarrow{\mathbf{h}};\overleftarrow{\mathbf{h}}]
\end{split}
\end{equation*}

This vector not only encodes the phonetic and phonological features, but it also captures some morphological patterns at the character level based on the IPA representations. Then, we concatenate this vector with the word and POS tag representations: $\mathbf{a} = [\mathbf{x}_t; \mathbf{p}_t; \mathbf{h}_{t}]$. We feed this representation to another bidirectional LSTM network \citep{dyer:2015acl}, similar to the BLSTM described for the character level. The bidirectional LSTM generates a word-level representation that accounts for the context in the sentence using semantics, syntax, phonetics and phonological aspects. We feed this representation to a fully-connected layer:
\begin{gather} \label{common_layer}
\mathbf{r}_i = ~\textrm{BLSTM}(\{\mathbf{a}_1, \mathbf{a}_2, ... \mathbf{a}_n\}) \\
\mathbf{z}_i = ~\textrm{ReLU}(\mathbf{W}_a \mathbf{r}_i + \mathbf{b})
\end{gather}
At this point, both models share the same definition. From here, we describe the multitask learning characteristics for each model separately. 

\begin{figure}
\centering
\includegraphics[width=7cm,height=6cm]{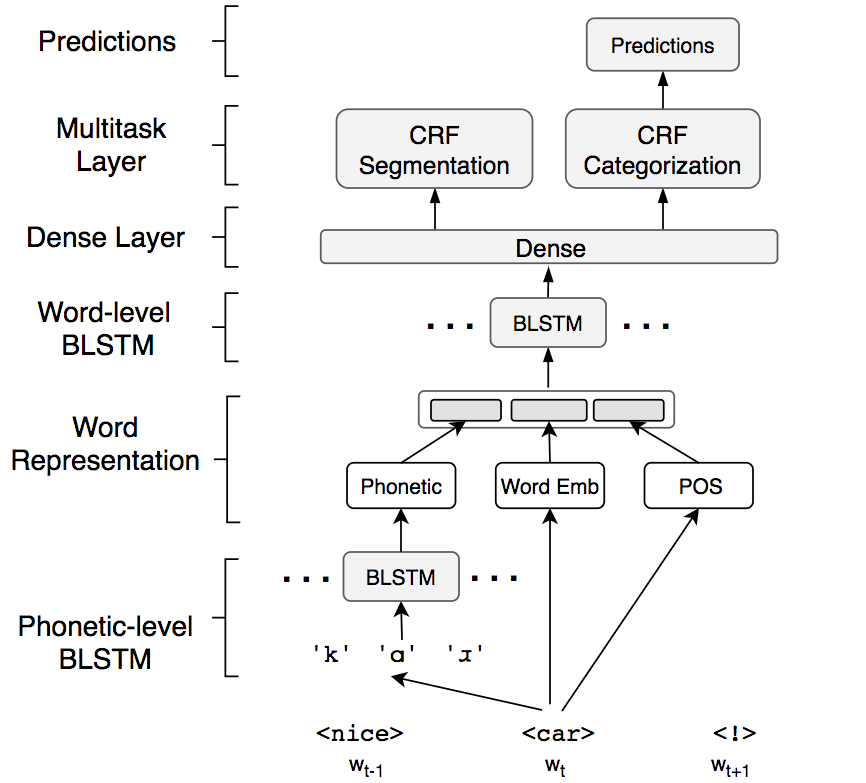}
\caption{ This is an end-to-end system that uses the CRF loss function as the objective function of the network. It also uses multitask learning on the output layer.}
\label{f:system1}
\end{figure}

\noindent \textbf{End-to-end model}. For the end-to-end network (see Figure \ref{f:system1}), we define an output layer based on two Conditional Random Fields \citep{Lafferty:2001:CRF:645530.655813}, each assigned to one of the tasks. The idea of adding a CRF to the model is to capture the relation of the output probabilities of the network with respect to the whole sequence. This means that the CRFs will maximize the log-likelihood of the entire sequence, which allows the model to learn very specific constraints from the data (e.g., a label \textit{I-location} cannot be followed by \textit{I-person}). Following \citet{MaAndHovy:16}, we formalize the definition of the CRF as follows: let $\mathbf{y} = [y_1, y_2, ..., y_n]$ be the labels for a sequence $\mathbf{x}$, where $y_i$ represents the $i^{th}$ label of the $\mathbf{x}_i$ token in the sentence. Next, we calculate the conditional probability of seeing $\mathbf{y}$ given the extracted features $\mathbf{z}$ from the network and the weights $\mathbf{W}$ associated to the labels:
\begin{equation*} \label{eq:crf}
p(\mathbf{y} | \mathbf{z}; \mathbf{W}) = \frac{\exp(\mathbf{W}_y{\Phi(\mathbf{z}, \mathbf{y})})}{\sum_{\mathbf{y'}\in \mathbf{y}} \exp(\mathbf{W}_y{\Phi(\mathbf{z}, \mathbf{y'})})}
\end{equation*}

Where $\Phi$  is a feature function that codifies the interactions between consecutive labels, $y_t$ and $y_{t+1}$, as well as the interactions between labels and words, represented by $z_t$. Then, the objective function for one CRF is defined by the maximum log-likelihood of this probability. However, we are running two CRFs as the objective function:
\begin{equation*} \label{eq:lstm_cell}
\begin{split}
\mathcal{L}_1(\mathbf{z}, \mathbf{W}) = &\log p(\mathbf{y}_{seg}| \mathbf{z}; \mathbf{W}) \\
\mathcal{L}_2(\mathbf{z}, \mathbf{W}) = &\log p(\mathbf{y}_{cat}| \mathbf{z}; \mathbf{W}) \\
L(\mathbf{z}, \mathbf{W}) = &~\alpha \mathcal{L}_1(\mathbf{z}, \mathbf{W}) + \mathcal{L}_2(\mathbf{z}, \mathbf{W}) \\
\end{split}
\end{equation*}

Where $\mathcal{L}_1$ is the loss function of the segmentation task with labels $\mathbf{y}_{seg}$. Similarly, $\mathcal{L}_2$ is the loss function of the categorization task with labels $\mathbf{y}_{cat}$. $L$ is the loss function that accounts for both tasks, where the segmentation task is weighted by an $\alpha$ scalar.

\noindent\textbf{Stacked model}. For this model, we use a multitask network as a feature extractor whose loss function is defined as a categorical cross entropy (see Figure \ref{f:system2}). We apply a softmax activation function to produce the probability distribution over the labels, and then we calculate the loss as follows:
\begin{equation*} \label{eq:xentropy}
\begin{split}
H_1(\mathbf{y}, \mathbf{z}) = & -\sum_{\mathbf{z}_i} \mathbf{y} \log(softmax(\mathbf{W}_{seg} \mathbf{z}_i + \mathbf{b})) \\
H_2(\mathbf{y}, \mathbf{z}) = & -\sum_{\mathbf{z}_i} \mathbf{y} \log(softmax(\mathbf{W}_{cat} \mathbf{z}_i + \mathbf{b})) \\
L(\mathbf{y}, \mathbf{z}) = & ~\alpha H_1(\mathbf{z}, \mathbf{W}_{seg}) + H_2(\mathbf{z}, \mathbf{W}_{cat}) \\
\end{split}
\end{equation*}
After training the multitask network, we take the activation outputs from Equation 2. These vectors are used as features to train a Conditional Random Fields classifier. The definition of the CRF is the same as the one described for the end-to-end network.

\subsection{Implementation details}

We have performed a very simple preprocessing on the data, which consists in replacing URLs, emojis, tags, and numbers with predefined tokens. Additionally, the vocabulary of the pretrained word embeddings was not sufficient to cover all the words in the WNUT dataset (i.e., training, validation, and testing sets have OOV words). We handled this situation using the Facebook library FastText \citep{bojanowski2016enriching}. 
This library can produce an embedding vector from the subword level of the word (i.e., ngrams). The advantage of FastText over other embedding learning algorithms is that we can still extract useful embeddings for OOV words from their subword embeddings. For instance, if there is a missing letter in one word, the subword-level vector will be reasonably close to the vector of the correct spelling.

The models have been trained using weighted classes, which forces the models to pay more attention to the labels that are less frequent. This is a very important step since the NE datasets usually show a skewed distribution, where the NE tokens represent approximately 10\% of the entire corpus. Although weighting classes improves the recall of the model, we tried to be sensitive to this aspect as the model can be forced to predict entities even in cases where there are none. The weights were experimentally defined, keeping the same distribution but decreasing the loss on non-entity tokens. 

Additionally, we defined our models using the following hyperparameters: the phonetic and phonological BLSTM at the character level uses 64 units per direction, which adds up to 128 units. Similarly, the word level BLSTM uses 100 units per direction, which accounts for a total of 200 units. The fully-connected layer has 100 neurons, and it uses a Rectified Linear Unit (ReLU) activation function. We also use a dropout operation before and after each BLSTM component. This forces the networks to find different paths to predict the data, which ultimately improves the generalization capabilities (i.e., they do not rely on a single path for certain inputs). The dropout value is 0.5. For the stacked model we use the Adam optimizer \citep{DBLP:journals/corr/KingmaB14} with a learning rate of 0.001.

\section{Datasets}
\label{sec:dataset}
\begin{table}
\centering
\begin{tabular}{llll}\\\hline
Corpus & Dataset & Classes & \% Unique \\\hline
\multirow{3}{*}{CoNLL 2003} & Train & 4 & 26\% \\  
 & Dev  &4 & 40\% \\  
 & Test  &4 & 41\% \\  \hline
\multirow{3}{*}{WNUT 2017} & Train & 6 & 75\% \\ 
 & Dev  & 6 & 85\% \\ 
 & Test & 6 & 80\% \\ \hline
\end{tabular}
\caption{Percentage of unique NEs in two benchmark datasets, the one from CoNLL 2003 and the one used in the 2017 shared task held by the WNUT workshop.}\label{t:stats_unique}
\end{table}

Social media (SM) captures the fast evolving behavior of the language, and, as its influence in society grows, SM platforms play an important role in language understanding. We focus this work on the WNUT 2017 dataset for NER \citep{DerczynskiEtAl:17}. This dataset covers multiple SM platforms and suits perfectly the purpose of this work. Table \ref{t:wnut17_dist} shows the distribution of the dataset and its classes. The training set uses tweets, whereas the development set is based on YouTube comments. The testing set combines content from Reddit and StackExchange. The cross domain nature of the dataset establishes an additional challenge to the task. For instance, besides the particularities of the domains (e.g., length of the sentences, domain-specific expressions such as hashtags, emojis and others), the users tend to address different topics on each of the SM domains with different levels of relaxed language and style~\citep{Ritter:2011:NER:2145432.2145595, StraussEtAl:16, DerczynskiEtAl:17}. Moreover, the predominant factors in those SM environments are the emerging and rare entities. As stated by \citet{DerczynskiEtAl:17}, \textit{emerging} describes the entity instances that started to appear in context recently (e.g., a movie title released a year ago), whereas \textit{rare} depicts the entities that appear less than certain number of times. It is worth noting that this dataset presents a great challenge to systems that rely on external resources due to the rare and emerging properties. 

We also consider the CoNLL 2003 dataset \citep{TjongKimSang-DeMeulder:03} as it has been used as the standard dataset for NER benchmarks. However, we emphasize that both datasets present significantly different challenges and, thus, some relevant aspects in CoNLL 2003 may not be that relevant in the WNUT 2017 dataset. For example, capitalization is a crucial feature in newswire text, but it is less important in SM data since users tend to arbitrarily alter the character casing. Moreover, the target classes on the WNUT 2017 dataset cover the CoNLL 2003 classes plus fine-grained classes such as \textit{creative-work} (e.g., movie titles, T.V. shows, etc.), \textit{group} (e.g., sports teams, music bands, etc.), and \textit{product}. The additional classes are more heterogeneous, and thus, it makes the task more difficult to generalize. Furthermore, Table \ref{t:stats_unique} shows the percentage of unique tokens of the WNUT 2017 dataset, which certainly shows a great diversity compared to the CoNLL 2003 dataset.

\begin{table}
\centering
\begin{tabular}{llll}
\hline
Statistics		& Train		& Dev		& Test \\\hline
Posts			& 3,395		& 1,009		& 1,287 \\
Tokens			& 62,729	& 15,733	& 23,394 \\
NE tokens		& 3,160		& 1,250		& 1,589 \\
NE tokens (\%)	& 5.04		& 7.95		& 6.79 \\\hline
\end{tabular}
\caption{General statistics of the WNUT 2017 dataset. It is worth noting that the NE tokens account for less than 10\% on any dataset, which shows the inherent skewness of the task. }
\label{t:wnut17_general_stats}
\end{table}

\begin{table}
\centering
\begin{tabular}{llll}
\hline
Classes			& Train	& Dev	& Test \\\hline
person			& 995	& 46	& 532 \\
location		& 793	& 238	& 188 \\
group			& 414	& 64	& 202 \\
creative-work	& 346	& 107	& 331 \\
product			& 345	& 586	& 250 \\
corporation		& 267	& 209	& 86 \\\hline
TOTAL			& 3,160	& 1,250	& 1,589 \\\hline
\end{tabular}
\caption{Classes and their frequency distribution on the WNUT 2017 dataset.}
\label{t:wnut17_dist}
\end{table}

\section{Experiments and results}
\label{sec:experiments}

\begin{table*}[t!]
\centering
\begin{tabular}{llllllllll}
\hline
\multirow{2}{*}{Classes} &
\multicolumn{3}{c}{Precision (\%)} &
\multicolumn{3}{c}{Recall (\%)} &
\multicolumn{3}{c}{F1 (\%)} \\
				& Stacked	& E2E    & WNUT   & Stacked  & E2E    & WNUT   & Stacked & E2E    & WNUT \\\hline
corporation 	& 33.33		& 30.77  & 31.91  & 19.70    & 12.12  & 22.73  & 24.76   & 17.39  & 26.55 \\
creative-work 	& 50.00		& 55.56  & 36.67  & 14.79    & 10.56  &  7.75  & 22.83   & 17.75  & 12.79\\ 
group 			& 47.76		& 63.16  & 41.79  & 19.39    & 14.55  & 16.97  & 27.59   & 23.65  & 24.14\\ 
location 		& 62.20		& 78.12  & 56.92  & 52.67    & 50.00  & 49.33  & 57.04   & 60.98  & 52.86\\ 
person 			& 73.49		& 71.15  & 70.72  & 51.05    & 51.75  & 50.12  & 60.25   & 59.92  & 58.66 \\
product 		& 40.58		& 34.29  & 30.77  & 22.05    &  9.45  &  9.45  & 28.57   & 14.81  & 14.46 \\\hline
Overall  		& 61.06		& \textbf{66.67}  & 57.54  &\textbf{36.33} 	 & 32.99  & 32.90  & \textbf{45.55}   & 44.14  & 41.86 \\
\hline
\end{tabular} 
\caption{The class-level and overall results of our systems on the WNUT 2017 dataset. WNUT represents the winning system of the shared task (UH-RiTUAL), E2E is the end-to-end model, and Stacked shows the results of the stacked model. Both systems considerably outperform the state-of-the-art results. Between the end-to-end and the stacked models, the former gets better overall precision while the latter stands out on recall.}
\label{t:results_classes}
\end{table*}

\begin{table}
\centering
\begin{tabular}{lll} \hline
Model & F1 & Delta\\\hline
Stacked Model & \textbf{45.55} & \\
~~~- Multitask Learning 	& 44.76  & -0.79\\
~~~- Character phonetics 	& 43.83  & -0.93\\
~~~- Weighted classes 		& 41.25	 & -2.58\\
~~~- POS tag vectors 		& 40.15  & -1.10\\
~~~- FastText OOV vectors 	& 39.78  & -0.37\\
~~~- Pretrained embeddings 	& 12.72  & -27.06\\
\hline
\end{tabular}
\caption{We performed an ablation experiment on the stacked model. The results in the table are the average of the scores of three iterations.}
\label{t:ablation}
\end{table}

We mainly focus our experiments on the WNUT 2017 dataset. However, we consider relevant to compare our approach to the standard CoNLL 2003 dataset where current state-of-the-art systems are benchmarked. This section addresses the experiments and results of both datasets.

\subsection{WNUT 2017 experiments}

In this section we discuss the experiments of the proposed approaches. We compare our models and describe the contribution of each component of the stacked system. Additionally, we compare our results against the state of the art in the WNUT 2017 dataset.

\noindent \textbf{Stacked vs. end-to-end model}. Table \ref{t:results_classes} shows that the stacked system has a lower precision than the end-to-end model, but its recall is the highest. This means that the stacked model is slightly better at generalizing than the other models since it can detect a more diverse set of entities. The surface form F1 metric \citep{DerczynskiEtAl:17} supports that intuition as well. It assigns a better F1 score to the stacked system (43.90\%) than to the end-to-end model (42.79\%) because the former finds more \textit{rare} and \textit{emerging} entities than the latter. Moreover, Table \ref{t:results_classes} also shows that the precision of the end-to-end model is higher than the rest of the systems. This tends to capture the most frequent entities and leave behind the \textit{rare} ones, which explains the different behaviors between the precision and recall of both models.

\noindent \textbf{Stacked model}. The feature extractor contains a category task that can produce predictions of the test set. We explored predicting the final labels with the feature extractor and compared the results against the predictions of the CRF classifier. We noticed that the CRF always outperformed the network. For the best scores the feature extractor achieved 40.64\% whereas the CRF reached 45.55\%. This is consistent with previous research \citep{DBLP:journals/corr/LampleBSKD16,AguilarEtAl:17} in that the individual output probabilities of the network do not consider the whole sequence, and thus, a sequential algorithm such as a CRF can improve the results by learning global constraints (i.e., the \textit{B-person} cannot be followed by \textit{I-corporation}).

\noindent \textbf{Ablation experiment}. We explored the contribution of the features and different aspects of our models. For instance, we tried a BLSTM network using pretrained word embeddings only. The results of this model set our baseline on a 39.78\% F1-score (see Table \ref{t:ablation}). This score is considerably close to the state-of-the-art performance, but improvements beyond that are small. For instance, Table \ref{t:ablation} shows an ablation experiment using the stacked model. The ablation reveals that weighting the classes is the most influential factor, which accounts for a 2.58\% of F1 score improvement. This aligns with the fact that the data is highly skewed, and thus, the model should pay more attention to the less frequent classes. The second most important aspect is the POS tags, which enhance the results by 1.10\%. This improvement suggests that POS tags are important whether the dataset is from a noisy environment or not since other researchers have found positive effects by using this feature on formal text \citep{DBLP:journals/corr/HuangXY15}. Almost equally influential are the phonetic and phonological features that push the F1 score by 0.93\%. According to the ablation experiment, using phonetic and phonology along with the pretrained word embeddings and POS tags can reach an F1 measure of 41.81\%, which is a very similar result to the state-of-the-art score, but with a simpler and more suitable model for SM environments (i.e., without gazetteers or capitalization).

We explored the multitask learning aspect by empirically trying multiple combinations of auxiliary tasks. The best combination is the standard NER categorization along with the segmentation task. The segmentation slightly improves the binary task proposed by \citet{AguilarEtAl:17} by around 0.3\%. Additionally, trying the binarization, segmentation, and categorization tasks together drops the results by around 0.2\% with respect to the categorization paired with the binary task. Moreover, the ablation experiment shows that the multitask layer boosts the performance of the stacked model with 0.79\% of F1 score. 

For the OOV problem, we use FastText to provide vectors to 2,333 words (around 13\% of the vocabulary). However, the ablation experiment shows a small improvement, which suggests that those words did not substantially contribute to the meaning of the context. Another aspect that we explored was adding all the letters of the dataset to the character level of the stacked model without modifying the casing. Surprisingly, the models produced a slightly worse result (around -0.5\%). Our intuition is that the character aspects are already captured by the model with the phonetic (IPA) representation, and the arbitrary use of capitalization renders this information useless. It is also worth noting that having phonetics instead of a language-dependent alphabet allows the adaptability of this approach to other languages.

\noindent \textbf{State of the art comparison}. Table \ref{t:results_classes} shows that our end-to-end and stacked models significantly outperform the state-of-the-art score by 2.28\% and 3.69\% F1 points, respectively. In the case of the stacked system, the precision and recall outperform the winning system of the shared task (UH-RiTUAL) across all the classes. Moreover, even though the UH-RiTUAL system uses gazetteers, it only outperforms the recall of the end-to-end model on the \textit{corporation} class. These results can be explained by the entity diversity of the dataset, where the \textit{emerging} and \textit{rare} properties are difficult to capture with external resources.

\subsection{CoNLL 2003 evaluation}

We also benchmarked our approach on a standard CoNLL 2003 dataset for the NER task. The stacked model reached 89.01\% while the end-to-end model achieved 88.98\% on the F1 metric. Although the state-of-the-art performance is 91.21\% \citep{MaAndHovy:16}, our approach targets SM domains and, consequently, our models disregard some of the important aspects on formal text while still getting reasonable results. For instance, \citet{MaAndHovy:16} input the text to their model \textit{as is}, which indirectly introduce capitalization to the morphological analysis at the character level. This aspect becomes relevant in this dataset because entities are usually capitalized on formal text. As explained before, our models do not rely on capitalization because the characters are represented by the International Phonetic Alphabet, which does not differentiate between lower and upper cases.

\section{Analysis}
\label{sec:analysis}
\begin{table}
\begin{center}
\begin{tabular}{|l|l|} \hline 
No & \bf Predictions \\ \hline
\multirow{2}{*}1 & 
	Road and airport closure isolate \underline{\textbf{Srinagar}} \\ 
    & as avalanche risk remains high \\\hline
\multirow{4}{*}2 & 
	\underline{The \textbf{Defence Research Development}} \\ 
  & \underline{\textbf{Organisation}} ( \underline{\textbf{DRDO}} ) is working on \\
  & four projects to develop new technologies \\
  & for more accurate ... \\\hline
3 & Her name is \textbf{Scout} . \\\hline
\end{tabular}
\end{center}
\caption{Examples of the predictions of our stacked model in the Reddit domain of the WNUT 2017 dataset. The bold words are the gold labels, and the underlined words are the predictions of our model. The model matches the entity types of the labeled data.}
\label{t:prediction_examples}
\end{table}
Table \ref{t:prediction_examples} shows some predictions of our stacked model on the WNUT 2017 test set. In example number 1, the model is able to correctly label \textit{Srinagar} as \textit{person}, even though the model does not rely on gazetteers or capitalization. It is also important to mention that the word was not in the training or development set, which means that the network had to infer the entity purely from the context. Moreover, the second example shows that the model has problems to determine whether the article \textit{the} belongs to an NE or not. This is an ambiguous problem that even humans struggle with. This example also has a variation on spelling for the words \textit{Defence} and \textit{Organisation}. We suspect that the mitigation of OOV words using the FastText library helped in this case. Also, from the phonetic perspective, the model treated the word \textit{Defence} as if it was the word \textit{Defense} because both words map to the same IPA sequence, \textipa{/dɪfɛns/}. 
In the third case, the model is not able to identify the NE \textit{Scout}, even though the context makes it fairly easy. 

\section{Related work}
\label{sec:relatedwork}
In its former years, NER systems focused on newswire text, where the goal was to identify mainly three types of entities: \textit{person}, \textit{corporation}, and \textit{location}. These entity types were originally proposed in the 6th Message Understanding Conference (MUC-6) \cite{Grishman:1996:MUC:992628.992709}. In MUC-7, the majority of the systems were based on heavily hand-crafted features and manually elaborated rules \cite{M98-1018}. Some  years later, many researchers incorporated machine learning algorithms to their systems, but there was still a strong dependency on external resources and domain-specific features and rules \cite{TjongKimSang-DeMeulder:03}. In addition, the majority of the systems used Maximum Entropy \cite{Bender-etAl:2003:CONLL, Chieu-Ng:2003:CONLL, Curran:2003:LIN:1119176.1119200, Florian-etAl:2003:CONLL, Klein:2003:NER:1119176.1119204} and Hidden Markov Models \cite{Florian-etAl:2003:CONLL, Klein:2003:NER:1119176.1119204, Mayfield:2003:NER:1119176.1119205, Whitelaw:2003:NER:1119176.1119208}. Furthermore, \citet{McCallum:2003:ERN:1119176.1119206} used a CRF combined with web-augmented lexicons. The features were selected by hand-crafted rules and refined based on their relevance to the domain of the entities. Moreover, \citet{Nothman:2013:LMN:2405838.2405915} used Wikipedia resources to take advantage of structured data and reduce the human-annotated labels. In general, the results of the systems were reasonable for formal text, yet the scalability and the expensive detailed rules were not; their systems were difficult to maintain and adapt to other domains where different rules were needed.

Recently, NER has been focused on noisy data as a result of the growth in social media users. However, the limits of the previous systems dramatically affected the results on noisy domains. For instance, \citet{DBLP:journals/corr/DerczynskiM0EGTPB14} evaluated multiple NER tools in noisy environments: Stanford NER \cite{Finkel:2005:INI:1219840.1219885}, ANNIE \cite{cunningham-EtAl:2002:ACL}, among others. They reported that the majority of the tools were not capable of adapting to the noisy conditions showing a drop in performance of around 40\% on a F1-score metric. This motivated many researchers to solve the problem using different techniques. In 2015, \citet{baldwin-EtAl:2015:WNUT} organized a NER shared task at the 1st Workshop on Noisy User-generated Text (WNUT), where three of the participants used word embedding as features to train their traditional machine learning algorithms \citep{godin2015multimedia, toh-chen-su:2015:WNUT, cherry-guo-dai:2015:WNUT}. The shared task introduced noisy data as well as more difficult entity types to identify (e.g., tv show, product, sports team, movie, music artist, etc.). Notably, the WNUT 2016 and 2017 were predominated by neural network systems \citep{LimsopathamAndCollier:16, AguilarEtAl:17}.

Deep neural networks have proven to be effective for NER. The state-of-the-art and the most competitive architectures can be characterized by the use of recurrent neural networks \citep{DBLP:journals/corr/ChiuN15} combined with CRF \citep{DBLP:journals/corr/LampleBSKD16, MaAndHovy:16, peng-dredze:2016:P16-2, bharadwaj-EtAl:2016:EMNLP2016, AguilarEtAl:17}. Our work primarily focuses on social media data and explores more suitable variations and combinations of those models. The most important differences of our approach and previous works are i) the use of phonetics and phonology (articulatory) features at the character level to model SM noise, ii) consistent BLSTMs for character and word levels, iii) the segmentation and categorization tasks, iv) a multitask neural network that transfers the learning without using lexicons or gazetteers, and v) weighted classes to handle the inherent skewness of the datasets.

\section{Conclusions}
\label{sec:conclusion}
This paper proposed two models for NER on social media environments. The first one is a stacked model that uses a multitask BLSTM network as a feature extractor to transfer the learning to a CRF classifier. The second one is an end-to-end multitask BLSTM-CRF model whose output layer has a CRF per task. Both models improve the state-of-the-art results on the WNUT 2017 dataset, where the data comes from multiple SM domains (i.e., Twitter, YouTube, Reddit, and StackExchange). Instead of working on normalizing text, we designed representations that are robust to inherent properties of SM data: inconsistent spellings, diverse vocabulary, and flexible grammar. Considering that SM is a prevalent communication channel that constantly generates massive amounts of data, it is practical to design NLP tools to process this domain \textit{as is}. In this sense, we showed that the phonetic and phonological features are useful to capture sound-driven writing. This approach avoids the standard normalization process and boosts prediction performance. Furthermore, the use of multitask learning with segmentation and categorization is important to improve the results of the models. Finally, the weighted classes force the model to pay more attention on skewed datasets. We showed that these components can point to more suitable approaches for NER on social media data.
\bibliography{naaclhlt2018}

\begin{thebibliography}{}
\expandafter\ifx\csname natexlab\endcsname\relax\def\natexlab#1{#1}\fi

\bibitem[{Aguilar et~al.(2017)Aguilar, Maharjan, L\'{o}pez~Monroy, and
  Solorio}]{AguilarEtAl:17}
Gustavo Aguilar, Suraj Maharjan, Adrian~Pastor L\'{o}pez~Monroy, and Thamar
  Solorio. 2017.
\newblock \href{http://www.aclweb.org/anthology/W17-4419}{A multi-task approach
  for named entity recognition in social media data}.
\newblock In {\em Proceedings of the EMNLP 3rd Workshop on Noisy User-generated
  Text\/}. Association for Computational Linguistics, Copenhagen, Denmark,
  pages 148--153.
\newblock \url{http://www.aclweb.org/anthology/W17-4419}.

\bibitem[{Augenstein et~al.(2017)Augenstein, Derczynski, and
  Bontcheva}]{AugensteinEtAl:17}
Isabelle Augenstein, Leon Derczynski, and Kalina Bontcheva. 2017.
\newblock \href{http://arxiv.org/abs/1701.02877}{Generalisation in named entity
  recognition: {A} quantitative analysis}.
\newblock {\em CoRR\/} abs/1701.02877.
\newblock \url{http://arxiv.org/abs/1701.02877}.

\bibitem[{Bender et~al.(2003)Bender, Och, and Ney}]{Bender-etAl:2003:CONLL}
Oliver Bender, Franz~Josef Och, and Hermann Ney. 2003.
\newblock \href{http://www.aclweb.org/anthology/W03-0420.pdf}{Maximum entropy
  models for named entity recognition}.
\newblock In Walter Daelemans and Miles Osborne, editors, {\em Proceedings of
  the Seventh Conference on Natural Language Learning at HLT-NAACL 2003\/}.
  pages 148--151.
\newblock \url{http://www.aclweb.org/anthology/W03-0420.pdf}.

\bibitem[{Bharadwaj et~al.(2016)Bharadwaj, Mortensen, Dyer, and
  Carbonell}]{bharadwaj-EtAl:2016:EMNLP2016}
Akash Bharadwaj, David Mortensen, Chris Dyer, and Jaime Carbonell. 2016.
\newblock \href{https://aclweb.org/anthology/D16-1153}{Phonologically aware
  neural model for named entity recognition in low resource transfer settings}.
\newblock In {\em Proceedings of the 2016 Conference on Empirical Methods in
  Natural Language Processing\/}. Association for Computational Linguistics,
  Austin, Texas, pages 1462--1472.
\newblock \url{https://aclweb.org/anthology/D16-1153}.

\bibitem[{Bojanowski et~al.(2016)Bojanowski, Grave, Joulin, and
  Mikolov}]{bojanowski2016enriching}
Piotr Bojanowski, Edouard Grave, Armand Joulin, and Tomas Mikolov. 2016.
\newblock Enriching word vectors with subword information.
\newblock {\em arXiv preprint arXiv:1607.04606\/} .

\bibitem[{Borthwick et~al.(1998)Borthwick, Sterling, Agichtein, and
  Grishman}]{M98-1018}
A.~Borthwick, J.~Sterling, E.~Agichtein, and R.~Grishman. 1998.
\newblock \href{http://www.aclweb.org/anthology/M98-1018}{Nyu: Description of
  the mene named entity system as used in muc-7}.
\newblock In {\em Seventh Message Understanding Conference (MUC-7): Proceedings
  of a Conference Held in Fairfax, Virginia, April 29 - May 1, 1998\/}.
\newblock \url{http://www.aclweb.org/anthology/M98-1018}.

\bibitem[{Cherry et~al.(2015)Cherry, Guo, and Dai}]{cherry-guo-dai:2015:WNUT}
Colin Cherry, Hongyu Guo, and Chengbi Dai. 2015.
\newblock \href{http://www.aclweb.org/anthology/W15-4307}{Nrc: Infused phrase
  vectors for named entity recognition in twitter}.
\newblock In {\em Proceedings of the Workshop on Noisy User-generated Text\/}.
  Association for Computational Linguistics, Beijing, China, pages 54--60.
\newblock \url{http://www.aclweb.org/anthology/W15-4307}.

\bibitem[{Chieu and Ng(2003{\natexlab{a}})}]{Chieu:2003:NER}
Hai~Leong Chieu and Hwee~Tou Ng. 2003{\natexlab{a}}.
\newblock \href{https://doi.org/10.3115/1119176.1119199}{Named entity
  recognition with a maximum entropy approach}.
\newblock In {\em Proceedings of the Seventh Conference on Natural Language
  Learning at HLT-NAACL 2003 - Volume 4\/}. Association for Computational
  Linguistics, Stroudsburg, PA, USA, CONLL '03, pages 160--163.
\newblock \url{https://doi.org/10.3115/1119176.1119199}.

\bibitem[{Chieu and Ng(2003{\natexlab{b}})}]{Chieu-Ng:2003:CONLL}
Hai~Leong Chieu and Hwee~Tou Ng. 2003{\natexlab{b}}.
\newblock \href{http://www.aclweb.org/anthology/W03-0423.pdf}{Named entity
  recognition with a maximum entropy approach}.
\newblock In Walter Daelemans and Miles Osborne, editors, {\em Proceedings of
  the Seventh Conference on Natural Language Learning at HLT-NAACL 2003\/}.
  pages 160--163.
\newblock \url{http://www.aclweb.org/anthology/W03-0423.pdf}.

\bibitem[{Chiu and Nichols(2016)}]{DBLP:journals/corr/ChiuN15}
Jason Chiu and Eric Nichols. 2016.
\newblock \href{https://transacl.org/ojs/index.php/tacl/article/view/792}{Named
  entity recognition with bidirectional lstm-cnns}.
\newblock {\em Transactions of the Association for Computational Linguistics\/}
  4:357--370.
\newblock \url{https://transacl.org/ojs/index.php/tacl/article/view/792}.

\bibitem[{Cunningham et~al.(2002)Cunningham, Maynard, Bontcheva, and
  Tablan}]{cunningham-EtAl:2002:ACL}
Hamish Cunningham, Diana Maynard, Kalina Bontcheva, and Valentin Tablan. 2002.
\newblock \href{https://doi.org/10.3115/1073083.1073112}{Gate: an architecture
  for development of robust hlt applications}.
\newblock In {\em Proceedings of 40th Annual Meeting of the Association for
  Computational Linguistics\/}. Association for Computational Linguistics,
  Philadelphia, Pennsylvania, USA, pages 168--175.
\newblock \url{https://doi.org/10.3115/1073083.1073112}.

\bibitem[{Curran and Clark(2003)}]{Curran:2003:LIN:1119176.1119200}
James~R. Curran and Stephen Clark. 2003.
\newblock \href{https://doi.org/10.3115/1119176.1119200}{Language independent
  ner using a maximum entropy tagger}.
\newblock In {\em Proceedings of the Seventh Conference on Natural Language
  Learning at HLT-NAACL 2003 - Volume 4\/}. Association for Computational
  Linguistics, Stroudsburg, PA, USA, CONLL '03, pages 164--167.
\newblock \url{https://doi.org/10.3115/1119176.1119200}.

\bibitem[{Derczynski et~al.(2014)Derczynski, Maynard, Rizzo, van Erp, Gorrell,
  Troncy, Petrak, and Bontcheva}]{DBLP:journals/corr/DerczynskiM0EGTPB14}
Leon Derczynski, Diana Maynard, Giuseppe Rizzo, Marieke van Erp, Genevieve
  Gorrell, Rapha{\"{e}}l Troncy, Johann Petrak, and Kalina Bontcheva. 2014.
\newblock \href{http://arxiv.org/abs/1410.7182}{Analysis of named entity
  recognition and linking for tweets}.
\newblock {\em CoRR\/} abs/1410.7182.
\newblock \url{http://arxiv.org/abs/1410.7182}.

\bibitem[{Derczynski et~al.(2017)Derczynski, Nichols, van Erp, and
  Limsopatham}]{DerczynskiEtAl:17}
Leon Derczynski, Eric Nichols, Marieke van Erp, and Nut Limsopatham. 2017.
\newblock \href{http://aclweb.org/anthology/W/W17/W17-4418.pdf}{{Results of the
  WNUT2017 Shared Task on Novel and Emerging Entity Recognition}}.
\newblock In {\em Proceedings of the 3rd Workshop on Noisy, User-generated Text
  (W-NUT) at EMNLP\/}. ACL.
\newblock \url{http://aclweb.org/anthology/W/W17/W17-4418.pdf}.

\bibitem[{Dyer et~al.(2015)Dyer, Ballesteros, Ling, Matthews, and
  Smith}]{dyer:2015acl}
Chris Dyer, Miguel Ballesteros, Wang Ling, Austin Matthews, and Noah~A. Smith.
  2015.
\newblock Transition-based dependeny parsing with stack long short-term memory.
\newblock In {\em Proceedings of the $53^{rd}$ Annual Meeting of the
  Association of Computational Linguistics and the $7^{th}$ International Joint
  Conference on Natural Language Processing of the Asian Federation of Natural
  Language Processing (ACL-IJCNLP 2015)\/}. ACL.

\bibitem[{Finkel et~al.(2005)Finkel, Grenager, and
  Manning}]{Finkel:2005:INI:1219840.1219885}
Jenny~Rose Finkel, Trond Grenager, and Christopher Manning. 2005.
\newblock \href{https://doi.org/10.3115/1219840.1219885}{Incorporating
  non-local information into information extraction systems by gibbs sampling}.
\newblock In {\em Proceedings of the 43rd Annual Meeting on Association for
  Computational Linguistics\/}. Association for Computational Linguistics,
  Stroudsburg, PA, USA, ACL '05, pages 363--370.
\newblock \url{https://doi.org/10.3115/1219840.1219885}.

\bibitem[{Florian et~al.(2003{\natexlab{a}})Florian, Ittycheriah, Jing, and
  Zhang}]{Florian:2003:NER}
Radu Florian, Abe Ittycheriah, Hongyan Jing, and Tong Zhang.
  2003{\natexlab{a}}.
\newblock \href{https://doi.org/10.3115/1119176.1119201}{Named entity
  recognition through classifier combination}.
\newblock In {\em Proceedings of the Seventh Conference on Natural Language
  Learning at HLT-NAACL 2003 - Volume 4\/}. Association for Computational
  Linguistics, Stroudsburg, PA, USA, CONLL '03, pages 168--171.
\newblock \url{https://doi.org/10.3115/1119176.1119201}.

\bibitem[{Florian et~al.(2003{\natexlab{b}})Florian, Ittycheriah, Jing, and
  Zhang}]{Florian-etAl:2003:CONLL}
Radu Florian, Abe Ittycheriah, Hongyan Jing, and Tong Zhang.
  2003{\natexlab{b}}.
\newblock \href{http://www.aclweb.org/anthology/W03-0425.pdf}{Named entity
  recognition through classifier combination}.
\newblock In Walter Daelemans and Miles Osborne, editors, {\em Proceedings of
  the Seventh Conference on Natural Language Learning at HLT-NAACL 2003\/}.
  pages 168--171.
\newblock \url{http://www.aclweb.org/anthology/W03-0425.pdf}.

\bibitem[{Godin et~al.(2015)Godin, Vandersmissen, De~Neve, and Van~de
  Walle}]{godin2015multimedia}
Fr\'{e}deric Godin, Baptist Vandersmissen, Wesley De~Neve, and Rik Van~de
  Walle. 2015.
\newblock \href{http://www.aclweb.org/anthology/W15-4322}{{Multimedia Lab $@$
  ACL WNUT NER Shared Task: Named Entity Recognition for Twitter Microposts
  using Distributed Word Representations}}.
\newblock In {\em Proceedings of the Workshop on Noisy User-generated Text\/}.
  Association for Computational Linguistics, Beijing, China, pages 146--153.
\newblock \url{http://www.aclweb.org/anthology/W15-4322}.

\bibitem[{Grishman and Sundheim(1996{\natexlab{a}})}]{GrishmanAndSundheim:96}
Ralph Grishman and Beth Sundheim. 1996{\natexlab{a}}.
\newblock Message understanding conference - 6: A brief history.
\newblock In {\em COLING-1996 Vol. 1\/}. pages 466--471.

\bibitem[{Grishman and
  Sundheim(1996{\natexlab{b}})}]{Grishman:1996:MUC:992628.992709}
Ralph Grishman and Beth Sundheim. 1996{\natexlab{b}}.
\newblock \href{https://doi.org/10.3115/992628.992709}{Message understanding
  conference-6: A brief history}.
\newblock In {\em Proceedings of the 16th Conference on Computational
  Linguistics - Volume 1\/}. Association for Computational Linguistics,
  Stroudsburg, PA, USA, COLING '96, pages 466--471.
\newblock \url{https://doi.org/10.3115/992628.992709}.

\bibitem[{Hochreiter and
  Schmidhuber(1997)}]{Hochreiter:1997:LSM:1246443.1246450}
Sepp Hochreiter and J\"{u}rgen Schmidhuber. 1997.
\newblock \href{https://doi.org/10.1162/neco.1997.9.8.1735}{Long short-term
  memory}.
\newblock {\em Neural Comput.\/} 9(8):1735--1780.
\newblock \url{https://doi.org/10.1162/neco.1997.9.8.1735}.

\bibitem[{Huang et~al.(2015)Huang, Xu, and Yu}]{DBLP:journals/corr/HuangXY15}
Zhiheng Huang, Wei Xu, and Kai Yu. 2015.
\newblock \href{http://arxiv.org/abs/1508.01991}{Bidirectional {LSTM-CRF}
  models for sequence tagging}.
\newblock {\em CoRR\/} abs/1508.01991.
\newblock \url{http://arxiv.org/abs/1508.01991}.

\bibitem[{Kingma and Ba(2014)}]{DBLP:journals/corr/KingmaB14}
Diederik~P. Kingma and Jimmy Ba. 2014.
\newblock \href{http://arxiv.org/abs/1412.6980}{Adam: {A} method for stochastic
  optimization}.
\newblock {\em CoRR\/} abs/1412.6980.
\newblock \url{http://arxiv.org/abs/1412.6980}.

\bibitem[{Klein et~al.(2003)Klein, Smarr, Nguyen, and
  Manning}]{Klein:2003:NER:1119176.1119204}
Dan Klein, Joseph Smarr, Huy Nguyen, and Christopher~D. Manning. 2003.
\newblock \href{https://doi.org/10.3115/1119176.1119204}{Named entity
  recognition with character-level models}.
\newblock In {\em Proceedings of the Seventh Conference on Natural Language
  Learning at HLT-NAACL 2003 - Volume 4\/}. Association for Computational
  Linguistics, Stroudsburg, PA, USA, CONLL '03, pages 180--183.
\newblock \url{https://doi.org/10.3115/1119176.1119204}.

\bibitem[{Lafferty et~al.(2001)Lafferty, McCallum, and
  Pereira}]{Lafferty:2001:CRF:645530.655813}
John~D. Lafferty, Andrew McCallum, and Fernando C.~N. Pereira. 2001.
\newblock \href{http://dl.acm.org/citation.cfm?id=645530.655813}{Conditional
  random fields: Probabilistic models for segmenting and labeling sequence
  data}.
\newblock In {\em Proceedings of the Eighteenth International Conference on
  Machine Learning\/}. Morgan Kaufmann Publishers Inc., San Francisco, CA, USA,
  ICML '01, pages 282--289.
\newblock \url{http://dl.acm.org/citation.cfm?id=645530.655813}.

\bibitem[{Lample et~al.(2016)Lample, Ballesteros, Subramanian, Kawakami, and
  Dyer}]{DBLP:journals/corr/LampleBSKD16}
Guillaume Lample, Miguel Ballesteros, Sandeep Subramanian, Kazuya Kawakami, and
  Chris Dyer. 2016.
\newblock \href{http://arxiv.org/abs/1603.01360}{Neural architectures for named
  entity recognition}.
\newblock {\em CoRR\/} abs/1603.01360.
\newblock \url{http://arxiv.org/abs/1603.01360}.

\bibitem[{Limsopatham and Collier(2016)}]{LimsopathamAndCollier:16}
Nut Limsopatham and Nigel Collier. 2016.
\newblock \href{http://aclweb.org/anthology/W16-3920}{Bidirectional lstm for
  named entity recognition in twitter messages}.
\newblock In {\em Proceedings of the 2nd Workshop on Noisy User-generated Text
  (WNUT)\/}. The COLING 2016 Organizing Committee, Osaka, Japan, pages
  145--152.
\newblock \url{http://aclweb.org/anthology/W16-3920}.

\bibitem[{Ma and Hovy(2016)}]{MaAndHovy:16}
Xuezhe Ma and Eduard Hovy. 2016.
\newblock \href{http://www.aclweb.org/anthology/P16-1101}{End-to-end sequence
  labeling via bi-directional lstm-cnns-crf}.
\newblock In {\em Proceedings of the 54th Annual Meeting of the Association for
  Computational Linguistics (Volume 1: Long Papers)\/}. Association for
  Computational Linguistics, Berlin, Germany, pages 1064--1074.
\newblock \url{http://www.aclweb.org/anthology/P16-1101}.

\bibitem[{Mayfield et~al.(2003)Mayfield, McNamee, and
  Piatko}]{Mayfield:2003:NER:1119176.1119205}
James Mayfield, Paul McNamee, and Christine Piatko. 2003.
\newblock \href{https://doi.org/10.3115/1119176.1119205}{Named entity
  recognition using hundreds of thousands of features}.
\newblock In {\em Proceedings of the Seventh Conference on Natural Language
  Learning at HLT-NAACL 2003 - Volume 4\/}. Association for Computational
  Linguistics, Stroudsburg, PA, USA, CONLL '03, pages 184--187.
\newblock \url{https://doi.org/10.3115/1119176.1119205}.

\bibitem[{McCallum and Li(2003)}]{McCallum:2003:ERN:1119176.1119206}
Andrew McCallum and Wei Li. 2003.
\newblock \href{https://doi.org/10.3115/1119176.1119206}{Early results for
  named entity recognition with conditional random fields, feature induction
  and web-enhanced lexicons}.
\newblock In {\em Proceedings of the Seventh Conference on Natural Language
  Learning at HLT-NAACL 2003 - Volume 4\/}. Association for Computational
  Linguistics, Stroudsburg, PA, USA, CONLL '03, pages 188--191.
\newblock \url{https://doi.org/10.3115/1119176.1119206}.

\bibitem[{Mortensen et~al.(2016)Mortensen, Littell, Bharadwaj, Goyal, Dyer, and
  Levin}]{mortensen-EtAl:2016:COLING}
David~R. Mortensen, Patrick Littell, Akash Bharadwaj, Kartik Goyal, Chris Dyer,
  and Lori Levin. 2016.
\newblock \href{http://aclweb.org/anthology/C16-1328}{Panphon: A resource for
  mapping ipa segments to articulatory feature vectors}.
\newblock In {\em Proceedings of COLING 2016, the 26th International Conference
  on Computational Linguistics: Technical Papers\/}. The COLING 2016 Organizing
  Committee, Osaka, Japan, pages 3475--3484.
\newblock \url{http://aclweb.org/anthology/C16-1328}.

\bibitem[{Nothman et~al.(2013)Nothman, Ringland, Radford, Murphy, and
  Curran}]{Nothman:2013:LMN:2405838.2405915}
Joel Nothman, Nicky Ringland, Will Radford, Tara Murphy, and James~R. Curran.
  2013.
\newblock \href{https://doi.org/10.1016/j.artint.2012.03.006}{Learning
  multilingual named entity recognition from wikipedia}.
\newblock {\em Artif. Intell.\/} 194:151--175.
\newblock \url{https://doi.org/10.1016/j.artint.2012.03.006}.

\bibitem[{Owoputi et~al.(2013)Owoputi, O'Connor, Dyer, Gimpel, Schneider, and
  Smith}]{owoputi2013improved}
Olutobi Owoputi, Brendan O'Connor, Chris Dyer, Kevin Gimpel, Nathan Schneider,
  and Noah~A. Smith. 2013.
\newblock \href{http://www.aclweb.org/anthology/N13-1039}{{Improved
  Part-of-Speech Tagging for Online Conversational Text with Word Clusters}}.
\newblock In {\em Proceedings of the 2013 Conference of the North American
  Chapter of the Association for Computational Linguistics: Human Language
  Technologies\/}. Association for Computational Linguistics, Atlanta, Georgia,
  pages 380--390.
\newblock \url{http://www.aclweb.org/anthology/N13-1039}.

\bibitem[{Peng and Dredze(2016)}]{peng-dredze:2016:P16-2}
Nanyun Peng and Mark Dredze. 2016.
\newblock \href{http://anthology.aclweb.org/P16-2025}{Improving named entity
  recognition for chinese social media with word segmentation representation
  learning}.
\newblock In {\em Proceedings of the 54th Annual Meeting of the Association for
  Computational Linguistics (Volume 2: Short Papers)\/}. Association for
  Computational Linguistics, Berlin, Germany, pages 149--155.
\newblock \url{http://anthology.aclweb.org/P16-2025}.

\bibitem[{Ritter et~al.(2011)Ritter, Clark, Mausam, and
  Etzioni}]{Ritter:2011:NER:2145432.2145595}
Alan Ritter, Sam Clark, Mausam, and Oren Etzioni. 2011.
\newblock \href{http://dl.acm.org/citation.cfm?id=2145432.2145595}{Named entity
  recognition in tweets: An experimental study}.
\newblock In {\em Proceedings of the Conference on Empirical Methods in Natural
  Language Processing\/}. Association for Computational Linguistics,
  Stroudsburg, PA, USA, EMNLP '11, pages 1524--1534.
\newblock \url{http://dl.acm.org/citation.cfm?id=2145432.2145595}.

\bibitem[{Strauss et~al.(2016)Strauss, Toma, Ritter, de~Marneffe, and
  Xu}]{StraussEtAl:16}
Benjamin Strauss, Bethany Toma, Alan Ritter, Marie-Catherine de~Marneffe, and
  Wei Xu. 2016.
\newblock \href{http://aclweb.org/anthology/W16-3919}{Results of the wnut16
  named entity recognition shared task}.
\newblock In {\em Proceedings of the 2nd Workshop on Noisy User-generated Text
  (WNUT)\/}. The COLING 2016 Organizing Committee, Osaka, Japan, pages
  138--144.
\newblock \url{http://aclweb.org/anthology/W16-3919}.

\bibitem[{Tjong Kim~Sang and De~Meulder(2003)}]{TjongKimSang-DeMeulder:03}
Erik~F. Tjong Kim~Sang and Fien De~Meulder. 2003.
\newblock \href{http://www.aclweb.org/anthology/W03-0419.pdf}{Introduction to
  the conll-2003 shared task: Language-independent named entity recognition}.
\newblock In Walter Daelemans and Miles Osborne, editors, {\em Proceedings of
  the Seventh Conference on Natural Language Learning at HLT-NAACL 2003\/}.
  pages 142--147.
\newblock \url{http://www.aclweb.org/anthology/W03-0419.pdf}.

\bibitem[{Toh et~al.(2015)Toh, Chen, and Su}]{toh-chen-su:2015:WNUT}
Zhiqiang Toh, Bin Chen, and Jian Su. 2015.
\newblock \href{http://www.aclweb.org/anthology/W15-4321}{Improving twitter
  named entity recognition using word representations}.
\newblock In {\em Proceedings of the Workshop on Noisy User-generated Text\/}.
  Association for Computational Linguistics, Beijing, China, pages 141--145.
\newblock \url{http://www.aclweb.org/anthology/W15-4321}.

\bibitem[{Whitelaw and Patrick(2003)}]{Whitelaw:2003:NER:1119176.1119208}
Casey Whitelaw and Jon Patrick. 2003.
\newblock \href{https://doi.org/10.3115/1119176.1119208}{Named entity
  recognition using a character-based probabilistic approach}.
\newblock In {\em Proceedings of the Seventh Conference on Natural Language
  Learning at HLT-NAACL 2003 - Volume 4\/}. Association for Computational
  Linguistics, Stroudsburg, PA, USA, CONLL '03, pages 196--199.
\newblock \url{https://doi.org/10.3115/1119176.1119208}.

\bibitem[{Yang et~al.(2016)Yang, Salakhutdinov, and
  Cohen}]{DBLP:journals/corr/YangSC16}
Zhilin Yang, Ruslan Salakhutdinov, and William~W. Cohen. 2016.
\newblock \href{http://arxiv.org/abs/1603.06270}{Multi-task cross-lingual
  sequence tagging from scratch}.
\newblock {\em CoRR\/} abs/1603.06270.
\newblock \url{http://arxiv.org/abs/1603.06270}.

\end{thebibliography}
\bibliographystyle{acl_natbib}

\end{document}